\documentclass[letterpaper, 10 pt, conference]{ieeeconf}  % Comment this line out if you need a4paper

\usepackage{epsfig}
\usepackage{eso-pic}
\usepackage{xspace}
\usepackage{graphicx}
\usepackage{amsmath}
\usepackage{amssymb}
\usepackage[font=small]{caption}
\usepackage{balance}

\usepackage{bbm}
\usepackage{booktabs}
\usepackage{xcolor}
\usepackage{enumerate}
\let\labelindent\relax
\usepackage{enumitem}
\usepackage{subcaption}
\usepackage{tabulary,multirow,overpic,xcolor,subfloat}
\usepackage[pagebackref=false,breaklinks=true,letterpaper=true,colorlinks=true,bookmarks=false]{hyperref}
\usepackage[noadjust]{cite}
\usepackage{listings}
\usepackage[symbol]{footmisc}

\definecolor{codegreen}{rgb}{0,0.6,0}
\definecolor{codegray}{rgb}{0.5,0.5,0.5}
\definecolor{codepurple}{rgb}{0.58,0,0.82}
\definecolor{backcolour}{rgb}{0.95,0.95,0.92}

\lstdefinestyle{mystyle}{
    backgroundcolor=\color{backcolour},   
    commentstyle=\color{codegreen},
    keywordstyle=\color{magenta},
    numberstyle=\tiny\color{codegray},
    stringstyle=\color{codepurple},
    basicstyle=\ttfamily\footnotesize,
    breakatwhitespace=false,         
    breaklines=true,                 
    captionpos=b,                    
    keepspaces=true,                 
    numbers=left,                    
    numbersep=5pt,                  
    showspaces=false,                
    showstringspaces=false,
    showtabs=false,                  
    tabsize=2
}

\newcommand{\fref}[1]{Fig.~\ref{#1}}

\IEEEoverridecommandlockouts                              % This command is only needed if 
\title{\LARGE \bf
PyPose v0.6: The Imperative Programming Interface for Robotics\vspace{-10pt}}

\author{\fontsize{9pt}{9pt}\selectfont Zitong Zhan$^1$, Xiangfu Li$^2$, Qihang Li$^1$, Haonan He$^3$, Abhinav Pandey$^2$, Haitao Xiao$^4$, Yangmengfei Xu$^5$, Xiangyu Chen$^1$,\\Kuan Xu$^6$, Kun Cao$^6$, Zhipeng Zhao$^1$, Zihan Wang$^3$, Huan Xu$^7$, Zihang Fang$^8$, Yutian Chen$^3$, Wentao Wang$^2$, Xu Fang$^6$,\\Yi Du$^1$, Tianhao Wu$^3$, Xiao Lin$^7$, Yuheng Qiu$^3$, Fan Yang$^9$, Jingnan Shi$^{10}$, Shaoshu Su$^1$, Yiren Lu$^1$, Taimeng Fu$^1$, Karthik Dantu$^{1}$,\\Jiajun Wu$^{11}$, Lihua Xie$^6$, Marco Hutter$^9$, Luca Carlone$^{10}$, Sebastian Scherer$^3$, Daning Huang$^2$, Yaoyu Hu$^3$, Junyi Geng$^2$, Chen Wang$^{1}$\\
\url{https://pypose.org}
\vspace{-15pt}
\thanks{Corresponding Email: {\tt\small admin@pypose.org}}
\thanks{$^{1}$State University of New York at Buffalo, Buffalo, NY 14260, USA.}
\thanks{$^{2}$Pennsylvania State University, University Park, PA, 16802, USA.}
\thanks{$^{3}$Carnegie Mellon University, Pittsburgh, PA 15213, USA.}
\thanks{$^{4}$ZBL Co., Ltd., China.}
\thanks{$^{5}$University of Melbourne, Parkville VIC 3052, Australia.}
\thanks{$^{6}$Nanyang Technological University, Singapore 639798.}
\thanks{$^{7}$Georgia Institute of Technology, Atlanta, GA 30332, USA.}
\thanks{$^{8}$Northview High School, Johns Creek, GA 30097, USA.}
\thanks{$^{9}$ETH Zürich, 8092 Zürich, Switzerland.}
\thanks{$^{10}$Massachusetts Institute of Technology, Cambridge, MA 02139.}
\thanks{$^{11}$Stanford University, Stanford, CA 94305, USA.}
}

\begin{document}

\maketitle
\thispagestyle{empty}
\pagestyle{empty}

%%%%%%%%%%%%%%%%%%%%%%%%%%%%%%%%%%%%%%%%%%%%%%%%%%%%%%%%%%%%%%%%%%%%%%%%%%%%%%%%
\begin{abstract}
PyPose is an open-source library for robot learning. It combines a learning-based approach with physics-based optimization, which enables seamless end-to-end robot learning. It has been used in many tasks due to its meticulously designed application programming interface (API) and efficient implementation.  From its initial launch in early 2022, PyPose has experienced significant enhancements, incorporating a wide variety of new features into its platform. To satisfy the growing demand for understanding and utilizing the library and reduce the learning curve of new users, we present the fundamental design principle of the imperative programming interface, and showcase the flexible usage of diverse functionalities and modules using an extremely simple Dubins car example. We also demonstrate that the PyPose can be easily used to navigate a real quadruped robot with a few lines of code. 
\end{abstract}

%%%%%%%%%%%%%%%%%%%%%%%%%%%%%%%%%%%%%%%%%%%%%%%%%%%%%%%%%%%%%%%%%%%%%%%%%%%%%%%%
\section{Introduction}

PyPose is a Python-based, robotics-oriented, and open-source library designed for researchers and rapid prototyping \cite{wang2023pypose}.
Thanks to its adeptly crafted imperative programming approach, it offers swift customization for distinct applications.
PyPose has found use in diverse robotic applications, including odometry \cite{fu2023islam}, control \cite{pandey2023learning}, and planning \cite{yang2023iplanner}.
It aims to bridge the gap in robotic systems where deep learning-based methods and physics-based optimization often end up residing in separate modules implemented by different libraries, which can lead to suboptimal solutions. 

Since the last major release (v0.3\footnote[2]{A~history~of~PyPose~is~at~\href{https://github.com/pypose/pypose/releases}{https://github.com/pypose/pypose/releases}}), PyPose has experienced significant enhancements, incorporating a wide variety of new features into its platform. We have noticed that there is a growing demand for understanding and utilizing PyPose, especially regarding the rationale behind the application programming interface (API) design.
To reduce the learning curve of new users, this paper seeks to present the principle of the imperative interface by showcasing several examples covering diverse aspects of robotics, such as state estimation, planning, and control.
The contributions of this paper include
\begin{itemize}
    \item We present the design philosophy behind the imperative interface of the PyPose library. That is, the dynamic system class is a unified API for various functionalities such as state estimation, trajectory smoothing, and control. We show that each function can be done within a few lines of code, using a simple Dubins car example. 
    \item We demonstrate that the control loop can be extended to real-world robots seamlessly, thereby enhancing robots with the ability to utilize various out-of-the-box functionalities within PyPose.
\end{itemize}

\begin{figure}[t]
\centering
\vspace{8pt}
\includegraphics[width=1\linewidth]{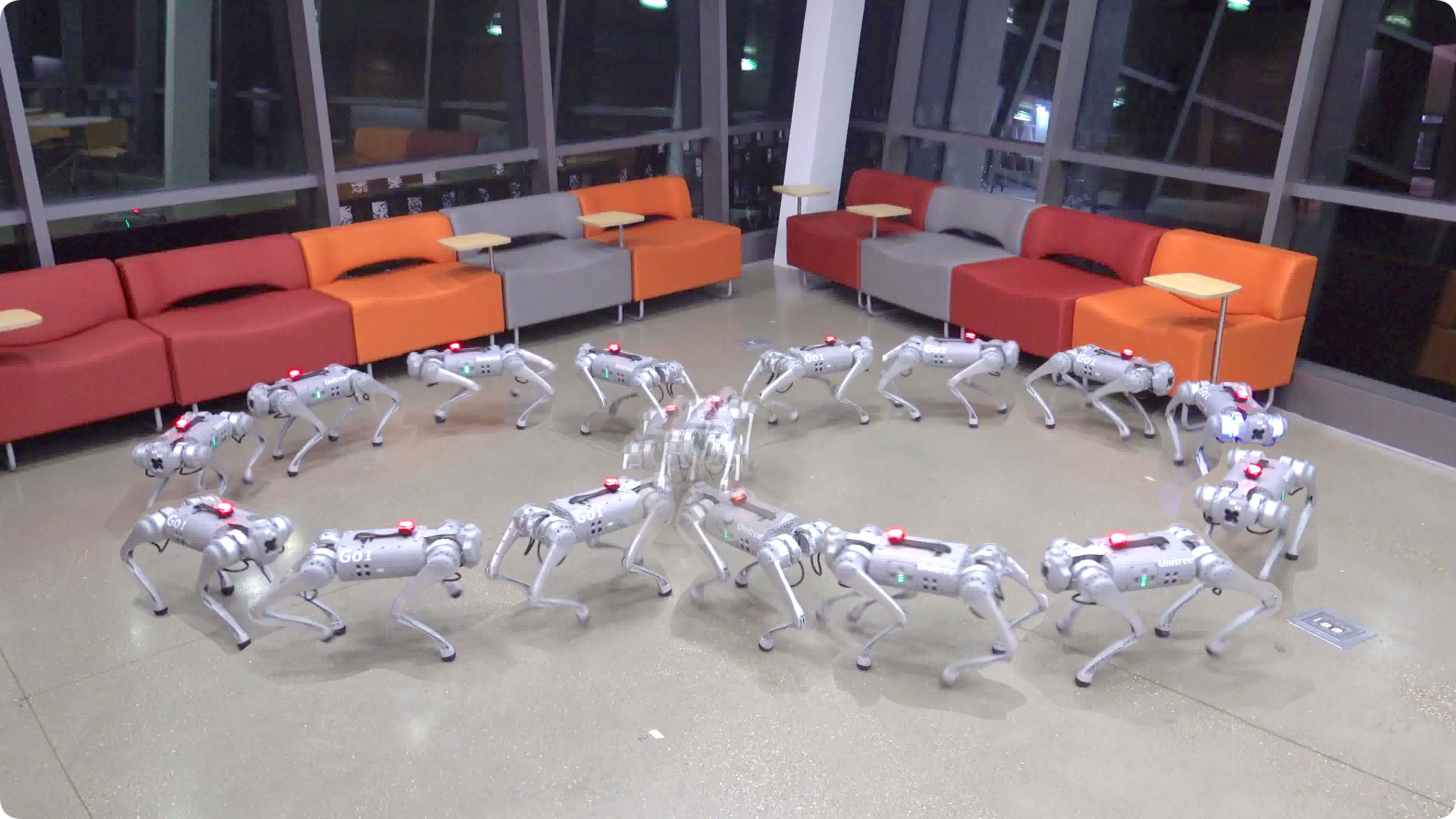}
\caption{With just a few lines of code in Python using PyPose, a quadruped robot can trace an $\infty$-shaped path.}
\label{fig:exp_real_robot}
\vspace{-12pt}
\end{figure}

The name ``PyPose" was inspired by an intriguing observation: the field of robotics can primarily be divided into four areas: perception, control, planning, and simultaneous localization and mapping (SLAM).
Essentially, robot control is about stabilizing its current pose, planning focuses on generating future poses, and SLAM deals with estimating the past and current poses.
To differentiate the library from PyTorch \cite{torch}, which excels in perception, and to emphasize its Python-centric approach and unique focus on robotics, the library was named ``PyPose".

\section{Related Work}

A comprehensive review was conducted in \cite{wang2023pypose}. In this study, we will compare some optimization libraries that are widely used in robotic applications.
Ipopt \cite{wachter2006implementation} is a C++-based solver targeting non-linear programming.
CasADi \cite{Andersson2018} is a non-linear optimization tool and implements forward mode algorithmic differentiation through a symbolic interface. It serves as a base tool for many software in planning and control.
However, both Ipopt and CasADi are not designed to work with robot learning methods.
While HILO-MPC \cite{pohlodek2022hilompc} based on CasADi attempts to integrate machine learning with optimal control models,  it does not provide easy-to-use native PyTorch support, and the control loop needs to be explicitly defined.
NeuroMANCER \cite{Neuromancer2023} is an open-source library that leverages PyTorch for differentiable programming, focusing on constrained optimization, dynamic system formulation, and parametric model-based optimal control. 
To the best of our knowledge, PyPose is a pioneering PyTorch-based library offering a comprehensive interface for robotics such as perception, SLAM, and control involving optimization.

\section{Imperative Programming Interface}

\begin{figure}[t]
\includegraphics[width=0.45\textwidth]{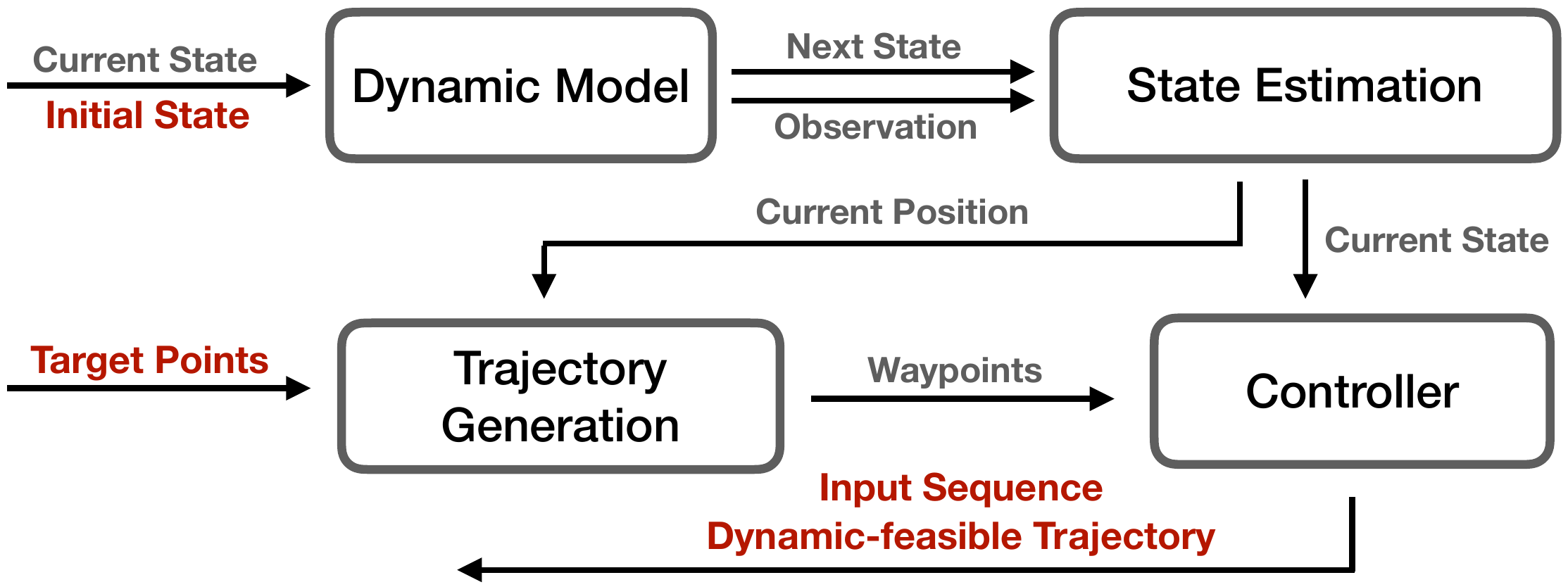}
\caption{The operational workflow
of the Dubins car system.}
\label{fig: logic}
\vspace{-12pt}
\end{figure}

Aside from the logical and modular design philosophies of PyPose, our principle of defining modules is to keep a general imperative interface that is compatible with a number of accessories. 
This section elucidates our design principles by focusing on four core modules: Dynamic Systems, State Estimation, Trajectory Interpolation, and Control. To illustrate the seamless integration and utility of these modules, we present a case study where a Dubins car is programmed to navigate an  $\infty$-like trajectory. The operational workflow of the entire system is depicted in \fref{fig: logic}.

\subsection{Dynamic System}
Dynamics provides a framework for describing the behavior of systems as they evolve over time. Such behavior can be generally modeled through two fundamental equations representing prediction $\mathbf{f}$ and observation $\mathbf{g}$:
\begin{subequations}
    \begin{align}
    \mathbf{x}_{k+1} &= \mathbf{f}(\mathbf{x}_k, \mathbf{u}_k, t_k) + \mathbf{w}_k\\
    \mathbf{y}_{k} &= \mathbf{g}(\mathbf{x}_k, \mathbf{u}_k, t_k) + \mathbf{v}_k
    \end{align}
\end{subequations}
where $t_k$, $\mathbf{x}_k\in\mathbb{R}^N$, $\mathbf{u}_k\in\mathbb{R}^C$, $\mathbf{y}_k\in\mathbb{R}^M$ are the system time, states, inputs, and observations at the $k$-th time step, respectively. The terms $\mathbf{w}$ and $\mathbf{v}$ represent noise in the state transition and observation functions, respectively.

We design \texttt{System} as a parent class for all dynamics systems including linear time-invariant (\texttt{LTI}) \cite{LTI}, linear time-variant (\texttt{LTV}) \cite{LTV}, and non-linear system (\texttt{NLS}) \cite{NLS}.
They not only carry the core functionality of performing state transition and system observation, but also provide the interface enabling users to access the key properties, such as linearized system matrices (automatically used by nonlinear control modules), etc.
\texttt{System} module and its sub-classes feature two user-defined class methods \texttt{state\_transition} and \texttt{observation}. 
To utilize those dynamic modeling capabilities, users simply need to subclass one of the predefined dynamic system templates and implement their specific state transition and observation methods according to their needs.

With this architecture, the inherited \texttt{forward} method from \texttt{torch.nn.Module} is used to handle state transitions for discrete-time systems and advances the time step. This design streamlines the implementation of discrete-time systems, which are commonly used in robotics.

We next demonstrate the simplicity of defining a system with a \texttt{DubinsCar} model in \eqref{eq:bicycle}.  
Generally, the state is defined as a position and heading in 2D plane and the inputs $\mathbf{u} = [v, \varphi]$ are linear velocity and angular rate
\begin{subequations}\label{eq:bicycle}
    \begin{align}
 \theta_{k+1} &= \theta_{k}+\varphi \cdot \Delta t \\
            {i}_{k+1} &= i_k + v \cdot \cos{\theta_{k+1}} \cdot \Delta t\\
            {j}_{k+1} &= j_k + v \cdot \sin{\theta_{k+1}}\cdot \Delta t
    \end{align}
\end{subequations}
In practical terms, to avoid the complications of angle wrapping, we opt for using the trigonometric values $\cos(\theta)$ and $\sin(\theta)$ instead of $\theta$ itself, and the system state is in the form of $[i, j, \cos(\theta), \sin(\theta)]$. This approach mitigates the issues of periodicity and discontinuity in angular values.

Due to its non-linearity, the \texttt{DubinsCar} model can be defined as a subclass of \texttt{NLS}, where the users only need to define methods \texttt{state\_transition} and \texttt{observation}.
\begin{lstlisting}[language=Python]
class DubinsCar(pp.module.NLS):
    # A 2-D DubinsCar kinematic model.
    def __init__(self, dt=0.1):
        super().__init__()
        self.dt = dt

    def state_transition(self, x, u, t=None):
        v, phi = u[..., 0], u[..., 1]
        c, s = x[..., 2], x[..., 3]
        theta = torch.atan2(s, c) + phi * dt

        i = x[..., 0] + v * c * dt
        j = x[..., 1] + v * s * dt
        c, s = theta.cos(), theta.sin()
        return torch.stack([i, j, c, s], dim=-1)

    def observation(self, x, u, t):
        return x
\end{lstlisting}

In this example, we override $\mathbf{f}$ in \texttt{state\_transition} and $\mathbf{g}$ in \texttt{observation}.
The model takes the current state \texttt{x} and an input \texttt{u[i]} as arguments, advances one time step, and returns the next state and observation.

\begin{lstlisting}[language=Python]
car = DubinsCar()
for i in range(N - 1):
    x[i+1], y[i] = car(x[i], u[i])
\end{lstlisting}

\begin{figure}[t]
    \begin{subfigure}[b]{0.24\textwidth}
    \centering
    \includegraphics[width=\textwidth]{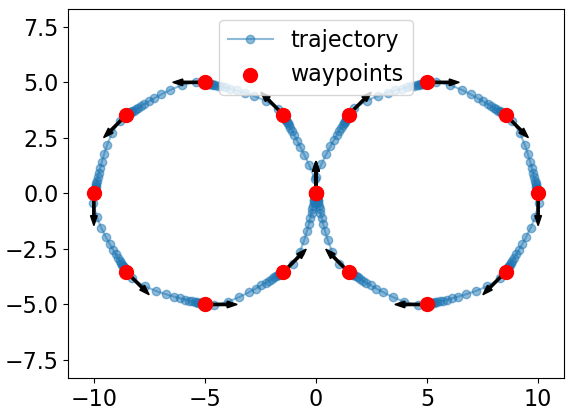}
    \caption{DubinCar trajectory in ideal case}
    \label{fig: Bicycle Trajectory}
    \end{subfigure}
    \begin{subfigure}[b]{0.24\textwidth}
    \centering
    \includegraphics[width=\textwidth]{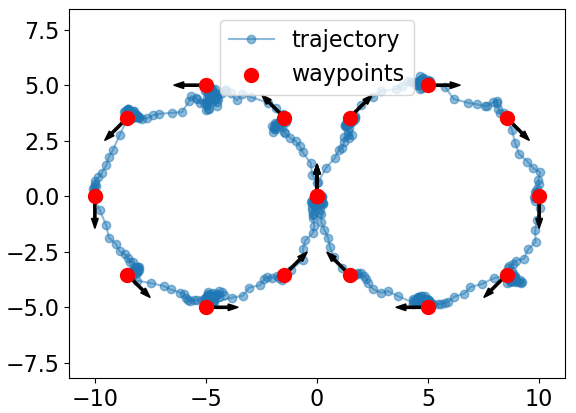}
    \caption{DubinCar trajectory with noise}
    \label{fig: Bicycle Trajectory with Noise}
    \end{subfigure}
    \caption{\texttt{DubinCar} system following waypoints.}
    \vspace{-12pt}
\end{figure}

\fref{fig: Bicycle Trajectory} shows an ideal \texttt{DubinsCar} model follows the $\infty$-trajectory without noise using an optimal input sequence calculated from the next section.

\begin{figure}[t]
    \begin{subfigure}[b]{0.24\textwidth}
    \centering
    \includegraphics[width=\textwidth]{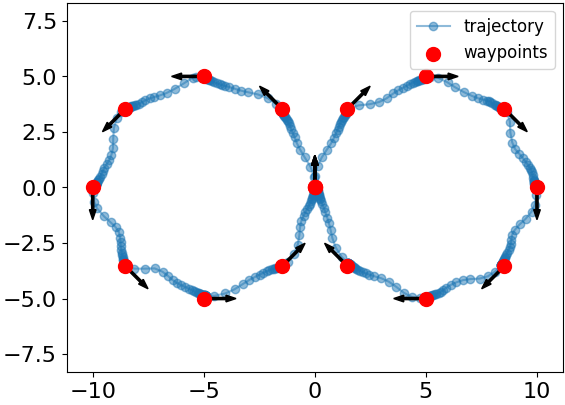}
    \caption{Raw noisy observation}  % 0.2045
    \label{fig: filter_none}
    \end{subfigure}
    \begin{subfigure}[b]{0.24\textwidth}
    \centering
    \includegraphics[width=\textwidth]{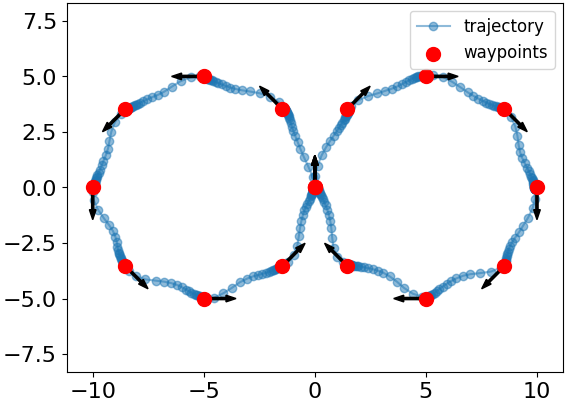}
    \caption{EKF estimation}  % 0.1803
    \label{fig: filter_ekf with all}
    \end{subfigure}
    \begin{subfigure}[b]{0.24\textwidth}
    \centering
    \includegraphics[width=\textwidth]{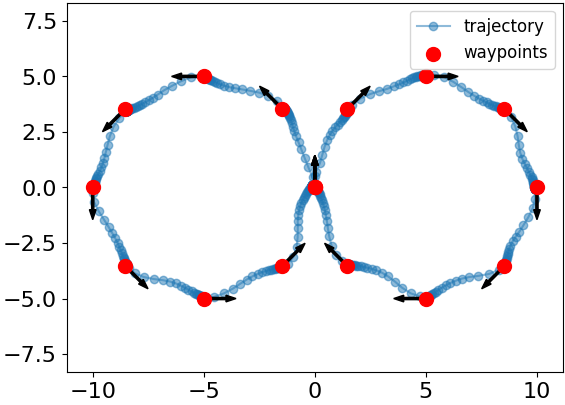}
    \caption{UKF estimation}  % 0.1802
    \label{fig: filter_ukf}
    \end{subfigure}
    \begin{subfigure}[b]{0.24\textwidth}
    \centering
    \includegraphics[width=\textwidth]{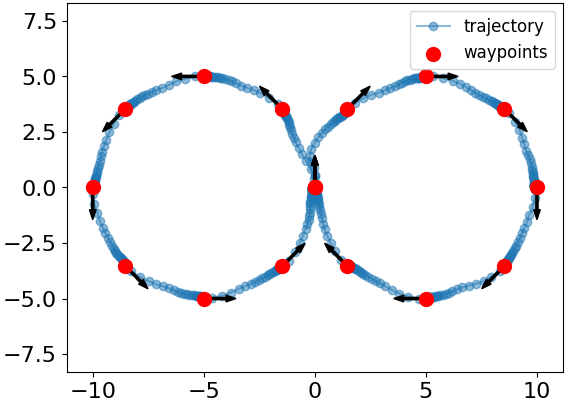}
    \caption{PF estimation}  % 0.0959
    \label{fig: filter_pf}
    \end{subfigure}
    \caption{MPC relying in different filters}
    \label{fig: all_filters}
    \vspace{-12pt}
\end{figure}
    
\subsection{Optimal Control Solvers} 
% LQR
% https://pypose.readthedocs.io/en/latest/generated/pypose.module.LQR.html

We next show that an optimal controller can be calculated on the fly based on the model's real-time state in one line of code.
PyPose provides several differentiable optimal controllers, including Linear Quadratic Regulator (\texttt{LQR}) \cite{LQR} and Model Predictive Control (\texttt{MPC}) \cite{MPC}.
LQR in PyPose can be used for both linear time-invariant (\texttt{LTI}) and linear time-varying (\texttt{LTV}) systems. 
Below we provide an example of MPC, controlling the \texttt{DubinsCar} model tracing waypoints on an $\infty$-shaped trajectory.
As illustrated in \fref{fig: Bicycle Trajectory with Noise}, the Dubins car visits each of the red waypoints, set as intermediate goal state, in order.
Its trajectory starts and ends at the center $(0, 0)$ location. 
The code snippet demonstrates MPC moving the Dubins car to traverse through all of the 17 waypoints (the start, end, and intermediate points overlap at the center location). 
Given the \texttt{target} parameter, the \texttt{MPC} computes the optimum input and iteratively moves the \texttt{DubinsCar} model until the current red target position is reached. Each blue connected dot represents the updated Dubins car state in one iteration.
\begin{lstlisting}[language=Python]
mpc = pp.module.MPC(car, Q, p, T)
\end{lstlisting}
where \texttt{Q} and \texttt{p} represent the weight matrices for the quadratic terms and the weight vectors for the linear terms at each time step, respectively. \texttt{T} is the time horizon on which MPC solves the optimization. Then an MPC object can be used as
\begin{lstlisting}[language=Python]
v = q * torch.randn(4) # observation noise
xt, u, cost = mpc(dt, y[i] + v, target)
\end{lstlisting}
where \texttt{dt} is the time step interval, \texttt{y[i]} is the observation from the previous iteration, \texttt{v} is the injected noice, and \texttt{target} is the goal state defined in the exact same format as the \texttt{DubinsCar} model's state; then the expected states \texttt{xt}, system inputs \texttt{u}, and \texttt{cost} along the time horizon will be outputted.
This example shows that the controller manages the robot's path effectively but the trajectory fluctuates in the presence of noise.
We next show that a state estimation module can be applied to further reduce the effect of noises.

\subsection{State Estimation}

PyPose provided the commonly used Bayesian filters \cite{simon2006optimal} including extend Kalman filter (\texttt{EKF}) \cite{EKF}, unscented Kalman filter (\texttt{UKF}) \cite{UKF}, and particle filter (\texttt{PF}) \cite{PF}.
Below we demonstrate the usage of \texttt{EKF} on the same \texttt{DubinsCar} system, and we hereby show that its usage only requires two additional lines of code. 
The first line of code defines the EKF module by wrapping up the Dubins car model. 
\begin{lstlisting}[language=Python]
ekf = pp.module.EKF(car)
\end{lstlisting}
Each time the system is propagated, the filter module takes the same format of input as the dynamics module. The second line of code is called immediately after the system is propagated with an observation produced. 
\begin{lstlisting}[language=Python]
X[i+1], P = ekf(X[i], y[i], u[i], P, Q, R)
\end{lstlisting}
where \texttt{X[i]} is the estimated state from the previous step and \texttt{y[i]} is the noisy observation; \texttt{P}, \texttt{Q}, and \texttt{R} are the state estimation covariance of the previous step, covariance of system transition model, and covariance of system observation model, respectively. The estimated observation could be used in MPC to generate more accurate control signals when observation noise is present. 
\fref{fig: all_filters} shows results produced by MPC when it takes either the raw noisy observation or the estimation from filters as input to follow the same trajectory.

\begin{figure}[t]
    \begin{subfigure}[b]{0.24\textwidth}
    \centering
    \includegraphics[width=\textwidth]{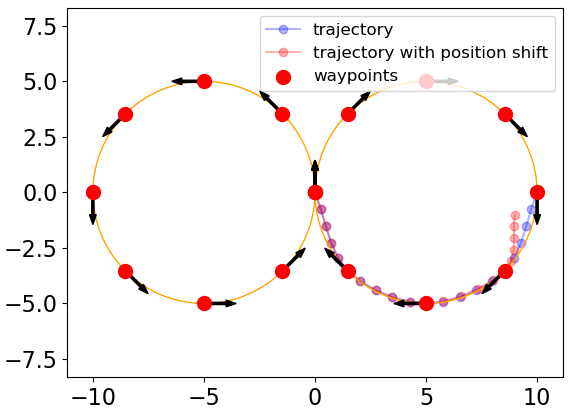}
    \caption{Local planned trajectory}
    \label{fig: local trajectory}
    \end{subfigure}
    \begin{subfigure}[b]{0.24\textwidth}
    \centering
    \includegraphics[width=\textwidth]{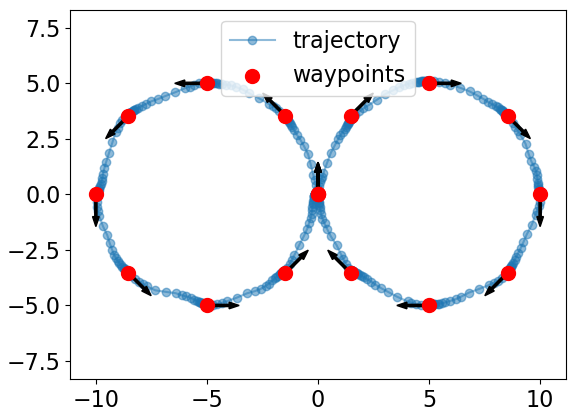}
    \caption{Trajectory with EKF and spline}
    \label{fig: Trajectory with all}
    \end{subfigure}
    \caption{\texttt{DubinCar} system with \texttt{chspline}}
    \vspace{-12pt}
\end{figure}

\subsection{Trajectory Interpolation}
PyPose provides trajectory interpolation algorithms, B-spline (\texttt{bspline}) \cite{bspline} and Cubic-Hermite Spline (\texttt{chspline}) \cite{chspline} as the guidance for trajectory smoothing.
The trajectory interpolation module serves as a valuable asset for robotic planning in uncertain environments, the dense smooth waypoints generated could guide the robot with a position shift back to the desired trajectory. Furthermore, the dense waypoints could decrease the computing cost of MPC and overshoot probability.
The \texttt{chspline} function in PyPose is user-friendly and straightforward to implement. By simply inputting a sequence of waypoints, the function generates an evenly distributed, smooth trajectory with a continuous first derivative. The code snippet below provides an example of how to employ the \texttt{chspline} function for online local trajectory planning with the \texttt{DubinsCar} model.

\begin{lstlisting}[language=Python]
points = pp.chspline(waypoints, interval=0.2)
\end{lstlisting}

\fref{fig: local trajectory} provides an example of local planning using the \texttt{spline} algorithm. The orange line serves as the ground truth trajectory, represented as two circles. The black arrows on the waypoints indicate the trajectory direction. The blue trajectory is generated in accordance with the planned path, while the red trajectory emerges when the robot deviates from this path due to an initial positional shift. Notably, the blue trajectory exhibits a high degree of alignment with the ground truth, attesting to the module's capability to generate optimal trajectories. 
Moreover, the red trajectory eventually merges smoothly with the blue one, illustrating the module's ability to produce smooth trajectories even when positional shifts occur with the Dubins car.

\fref{fig: Trajectory with all} depicts a trajectory generated by combining the \texttt{EKF} and \texttt{chspline} algorithms. When compared to the results solely from \texttt{EKF} as shown in \fref{fig: filter_ekf with all}, the inclusion of \texttt{chspline} noticeably smooths the trajectory and brings it closer to the desired $\infty$-shaped path. The algorithms not only generate optimal and smooth trajectories but also demonstrate resilience in accommodating positional shifts. 
These methods also support batch processing of waypoint sequences, enabling the generation of multiple trajectories simultaneously. This feature enhances computational efficiency and expedites the planning process.

\section{EXPERIMENTS}

\begin{figure}[t]
\centering
\includegraphics[height=3.6cm]{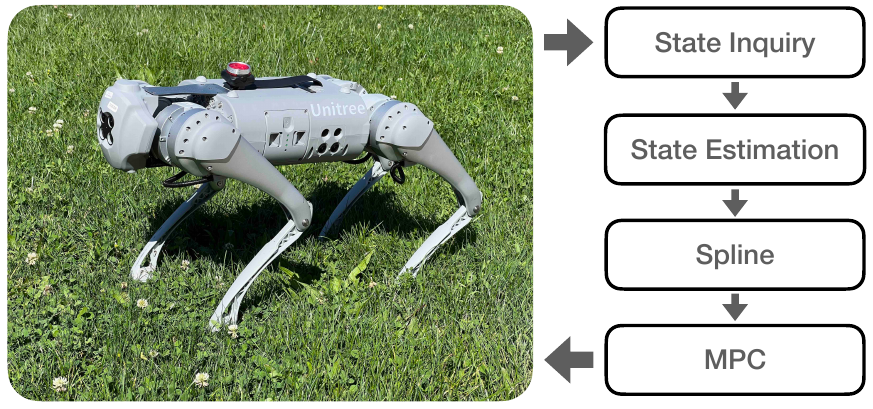}
\caption{Flow chart for the quadruped robot experiment.}
\label{fig:exp_logic}
\vspace{-12pt}
\end{figure}

This experiment aims to showcase the real-world applicability of PyPose by integrating it with a Unitree Go1 quadruped robot. The task involves adding the execution on a real robot into the PyPose implemented control loop, as indicated in Fig.~\ref{fig:exp_logic}, and navigating the robot through the waypoints on the $\infty$-shaped trajectory.
The assumption is that the robot has built-in localization producing reasonably accurate position estimation, and can be controlled through high-level motion commands. 
The controller on the robot is a cluster of single-board computers loaded with a Ubuntu desktop. 
The control loop could be executed on any remote computer within the same local network as the robot.

\begin{figure}[t]
\centering
\includegraphics[height=3.6cm]{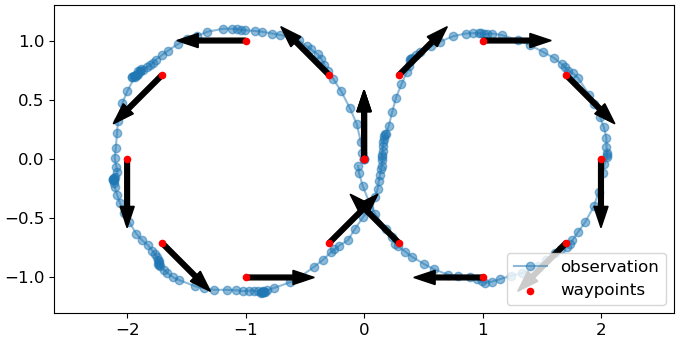}
\caption{The trajectory recorded on the quadruped robot in \fref{fig:exp_real_robot}.}
\label{fig:path}
\vspace{-12pt}
\end{figure}

In each iteration of the control loop, it inquires about the position and heading of the robot. Using the inquired pose as observation, the control loop sends a command to the robot based on optimal control from MPC, until the robot reaches a target.
Finally, the trajectory illustrated in \fref{fig:path} is recorded after the robot traverses through each of the target locations. Each of the blue connected dots represents an observation at a time step. Stacked snapshots of the real environment during execution are shown in \fref{fig:exp_real_robot}. 
This indicates that PyPose can effectively guide the robot along the desired path.

\section{CONCLUSIONS \& DISCUSSION}

In this paper, we highlighted the conciseness of the PyPose library's imperative interface. It offers a unified API for various functionalities, all achievable with a few lines of code. Our experiments showed that PyPose integrates seamlessly with a quadruped robot for navigation and equips robots with a myriad of ready-to-use features.
It is worth noting that all functions presented in this paper are fully differentiable, which makes an attractive starting point for the development of more advanced end-to-end robot learning systems. We expect that PyPose will inspire broader robotics research.

{
\balance
\small
\bibliographystyle{IEEEtran} 
\bibliography{egbib}
}

\end{document}